\pdfoutput=1
\documentclass[useAMS,usenatbib,referee]{biom}

\usepackage{algorithm}       
\usepackage{algpseudocode} 
\usepackage[colorlinks,citecolor=blue,urlcolor=blue]{hyperref}
\usepackage{lineno}

\usepackage{subfigure}
\usepackage{appendix}
\usepackage{graphicx}
\usepackage{bbm}
\usepackage{booktabs}
\usepackage{amsmath,amssymb}
\usepackage{multicol}
\usepackage[normalem]{ulem}

\def\beq{\begin{equation}}
\def\eeq{\end{equation}}
\def\beqr{\begin{eqnarray}}
\def\eeqr{\end{eqnarray}}
\def\beqrs{\begin{eqnarray*}}
\def\eeqrs{\end{eqnarray*}}
\def\bet{\begin{theorem}}
\def\eet{\end{theorem}}
\def\bel{\begin{lemma}}
\def\eel{\end{lemma}}
\def\bep{\begin{proposition}}
\def\eep{\end{proposition}}
\def\bg{\begin{figure}[tbph]\begin{center}}
\def\eg{\end{center}\end{figure}}

\def\bc{\begin{center}}
\def\ec{\end{center}}

\def\wt{\widetilde}
\def\wh{\widehat}

\def\mR{\mathbb{R}}
\def\mS{\mathcal S}

\def\mL{\mathcal L}
\def\mM\mathcal{M}

\def\1{\mbox{\boldmath $1$}}
\def\SEW{\widehat{\operatorname{SE}}(W)}

\def\vech{\operatorname{vech}}

\def\mM{\mathcal M}
\def\mF{\mathcal F}

\def\argmin{\mbox{argmin}}

\newcounter{proposition}
\renewcommand{\theproposition}{\arabic{proposition}}

\newenvironment{proposition}[1][]{%
    \refstepcounter{proposition}
    \noindent\textbf{Proposition \theproposition #1: }
    \setlength{\parindent}{0pt}
}{%
    \par
}

\def\semi{ \operatorname{semi} }

\def\msF{\mathsf{F}}
\def\SEF{ \wh{\operatorname{SE}}(\mF) }
\def\SEsF{ \wh{\operatorname{SE}}(\msF) }

\def\loc{\operatorname{loc}}

\def\mnem{\operatorname{mnem}}
\def\snem{\operatorname{snem}}
\newcommand{\Mean}{{\mathbb{E}}}

\title[Decentralized EM Algorithm]{Decentralized EM Algorithm for Gaussian Mixtures under \\ Data Heterogeneity and Partial Labeling}

\author{Xuetong Li$^{1}$, Shuyuan Wu$^{2,*}$\email{wushuyuan@mail.sufe.edu.cn}, Bin Du$^{3}$, and Hansheng Wang$^{4}$\\
$^{1}$School of Mathematics and Statistics, Xi'an Jiaotong University, Xi'an, China \\
$^{2}$School of Statistics and Data Science, Shanghai University of Finance and Economics, Shanghai, China \\
$^{3}$School of Mathematics and Statistics, Beijing Technology and Business University, Beijing, China \\
$^{4}$Guanghua School of Management, Peking University, Beijing, China}

\begin{document}



\pagerange{\pageref{firstpage}--\pageref{lastpage}} 
\volume{1}
\pubyear{2025}
\artmonth{March}

\doi{10.1111/j.1541-0420.2005.00454.x}

\label{firstpage}

\begin{abstract}
We systematically study several network-based Expectation–Maximization (EM) algorithms for the Gaussian mixture model  within decentralized federated learning (DFL). Our theoretical investigation reveals that directly extending the classic EM algorithm to DFL leads to a seriously biased estimator if the data are heterogeneously distributed across different sites. To address this issue, we introduce a momentum network EM (MNEM) algorithm, which integrates information from both current and historical estimators from previous DFL iterations. We further develop a semi-supervised MNEM (semi-MNEM) algorithm, which utilizes valuable information provided by partially labeled data. Rigorous theoretical analysis demonstrates that the MNEM estimator can achieve the same asymptotic efficiency as the whole sample estimator under appropriate regularity conditions, even if the data are heterogeneously distributed. Moreover, the semi-MNEM estimator significantly improves the convergence speed of the MNEM algorithm, even if different mixture components are poorly separated. Extensive simulations are conducted, and a widely used chest X-ray dataset is analyzed to demonstrate the finite-sample performance of the proposed methods.
\end{abstract}

\begin{keywords}
Distributed computing; Decentralized federated learning; Expectation-Maximization algorithm; Gaussian mixture model; Heterogeneity.
\end{keywords}


\maketitle


\section{Introduction}
\label{sec:intro}

\subsection{Background}

In recent years, lung diseases caused by different factors have led to a significant increase in mortality. 
For example, the Institute for Health Metrics and Evaluation recorded a cumulative death toll of 6.9 million due to COVID-19 by May 31, 2022 \citep{sachs2022lancet}.
Numerous deaths caused by COVID-19 were related to severe respiratory distress and significant cardiovascular failure due to substantial chest congestion.
As noted by \cite{chowdhury2020can} and \cite{Tawsifur2021},
the emergence of the COVID-19 pandemic has presented a challenge for clinical practitioners in distinguishing various lung diseases.
One of the most commonly used tools for diagnosing various lung diseases is the chest X-ray (CXR),
which is easily accessible and can offer both quick and precise diagnoses.
Therefore, early detection of lung diseases from CXR images is highly important \citep{yuan2021large}.

This study focuses on lung disease detection using CXR data via collaborative learning.
By collaborative learning, we mean training diagnostic models by integrating multi-site clinical information while keeping the CXR data local.
This approach leads to diagnostic models that are more robust and generalizable than those trained on a single site. Moreover, data privacy is also well protected. For collaborative learning purposes, various centralized federated learning (CFL) algorithms have been developed \citep{mcmahan2017communication}. 
They typically rely on a central server to update and then share model parameters/gradients across different sites. This leads to several inherent limitations \citep{liu2022decentralized}. 
First, it requires high network bandwidth since the central server must communicate with many sites simultaneously. Second, a CFL system is often very vulnerable in the sense that if the central server breaks down, the entire learning process stops. Third, a CFL system lacks strong privacy safeguards, as an attack to the central server might put all sites in danger.

To address the limitations of CFL, various decentralized federated learning (DFL) methods have been proposed \citep{shi2015extra,yuan2016convergence}. 
By decentralization, we mean that collective learning is conducted without a central server. All interactions should occur directly between individual sites.
The model parameters/gradients are then iteratively updated and shared between neighboring sites according to a prespecified communication network and protocol.
This leads to a more robust, secure, scalable, and collaborative learning system. 
Despite their usefulness, most DFL methods have been developed for supervised learning, for which many accurately annotated labels are needed.
In practice, however, obtaining accurate class labels for a large collection of CXR image data is time-consuming and labor-intensive, as it often requires experienced radiologists for every site \citep{zhou2023ensemble}.
Therefore, unsupervised and semi-supervised DFL is practically appealing.
Additionally, traditional algorithms struggle with data heterogeneity, which refers to the fact that data across different sites are not independently and identically distributed.
Data heterogeneity has been extensively documented in clinical research literature \citep{babar2024investigating,yan2023label}.
It stems from geographic differences, varied clinical practices, patient demographic diversity, and inconsistent medical device protocols; see \cite{ye2023heterogeneous} for a good summary. 
Therefore, how to handle heterogeneous data in DFL becomes an important problem.

\subsection{Related Work}

Our research is empirically motivated by CXR data analysis. Methods for detecting lung disease using CXR data have been rapidly developed. The key concept is to extract important features from CXR images for disease diagnosis. 
For example, \cite{xu2020radiomic} developed radiomic models to extract thousands of tumor-related features from CXR images. 
\cite{zhou2023ensemble} built deep learning models to learn features.
Various DFL methods have also been recently developed for CXR data analysis. 
For example, \cite{feki2021federated} proposed a DFL framework for multi-institution COVID-19 CXR screening.
\cite{bai2021advancing} constructed a DFL-based CT-COVID AI diagnostic initiative.
However, all these methods are supervised and thus require accurately annotated labels on a large scale.

Our research is also theoretically motivated by the algorithmic development of DFL. These DFL algorithms can be roughly classified into two categories. The first category contains various \textit{decentralized consensus optimization methods}, which enforce consensus among neighboring estimators \citep{shi2015extra}.
The second category contains various \textit{decentralized gradient descent methods} \citep{liu2022decentralized,gu2024statistical}. A key research question here is as follows: can the resulting DFL estimator achieve the same statistical efficiency as the whole sample estimator? Notable progress has been made by \cite{wu2023network} and \cite{gu2024statistical}.
They showed that the resulting estimator obtained from a carefully designed network gradient descent method might achieve the same statistical efficiency as that of a whole-sample estimator, as long as the network structure is fairly balanced and the learning rate is sufficiently small. However, all the aforementioned DFL algorithms are supervised learning methods, for which accurately annotated labels are needed. 
 
Our method is closely related to the Gaussian mixture model (GMM) since we use the GMM as an effective technique for unsupervised learning. In fact, the GMM has been one of the most widely used techniques for unsupervised learning. To estimate a GMM, the classic Expectation–Maximization (EM) algorithm has been developed \citep{wu1983convergence}, and its numerical convergence properties have been extensively studied \citep{chen1995optimal,JMLR:v25:23-1245}.
Distributed EM algorithms have also been developed for the GMM in FL \citep{nowak2003distributed,gu2008distributed}. 
These methods, however, do not consider data heterogeneity or poor separation. Unfortunately, both issues widely exist in practice. Moreover, the theoretical understanding of those algorithms from a statistical perspective is limited.

\subsection{Our Contributions}


Our first contribution is a novel DFL algorithm for unlabeled and partially labeled  CXR data analysis. 
To this end, we assume a standard GMM for all CXR data while accounting for data heterogeneity.
To estimate the GMM via DFL, we start with a straightforward combination of the standard EM algorithm and DFL, named the na\"ive network EM (NNEM) algorithm. 
Unfortunately, the NNEM estimator is consistent only under the ideal scenario of homogeneous data distribution and well-separated GMM components, yet remains statistically inefficient even in this case. 
To tackle heterogeneity, we develop a momentum network EM (MNEM) algorithm, which fuses current and historical estimators for effective cross-site information exchange.
This leads to an estimator achieving asymptotic efficiency equivalent to that of the whole-sample estimator.
However, the good properties of the MNEM algorithm rely on an appropriate separation condition that fails for poorly separated mixture components. 
To address this, we further propose a semi-supervised MNEM (semi-MNEM) algorithm, which leverages partial labels.
Theoretical analysis shows that semi-MNEM significantly boosts numerical convergence speed even if different mixture components are poorly separated.

Our second contribution concerns CXR data analysis. In this regard, we apply our (semi)-MNEM methods to a real-world, important CXR dataset.
This dataset is the COVID-19 dataset from \cite{chowdhury2020can} and \cite{Tawsifur2021}, which is one of the largest publicly accessible COVID-19-positive databases. 
This dataset has been used to enable data-intensive deep neural network models for disease pattern mining and disease correlation analysis.
Two competing methods are demonstrated for comparison purposes.
The results of our experiments show that the new methods outperform both methods.

The rest of the paper is organized as follows. Section \ref{sec:NNEM} presents the NNEM algorithm, and discusses its limitations. Section \ref{sec:methodology} introduces the proposed MNEM and semi-MNEM
algorithms along with their theoretical properties. Section \ref{sec:Numerical} presents numerical experiments based on both simulation datasets and a real-world dataset.
Finally, Section \ref{sec:discussion} concludes the paper with a brief
discussion. All the technical details are provided in the Appendix.

\section{A Na\"ive Network EM Algorithm} \label{sec:NNEM}

\subsection{GMM and the EM Algorithm}

Let $X_i = ( X_{ij} ) \in \mR^p$ be the $p$-dimensional feature collected from the $i$th subject with $1 \leq i \leq N$. To model the stochastic behavior of $X_i$, we adopt a widely used GMM. 
Specifically, assume that there exists a latent categorical random variable $Y_i \in \{1,2,\dots,K\}$ with $P(Y_i = k) = \alpha_k$, where $\alpha_k > 0$ for every $1 \leq k \leq K$ and $\sum_{k=1}^K \alpha_k = 1$. 
Given $Y_i = k$, assume $X_i$ follows a Gaussian distribution with mean $\mu_k \in \mR^p$ and variance $\Sigma_k \in \mR^{p \times p}$. The probability density function (PDF) of $X_i$ given $Y_i = k$ can be expressed as
$
\phi_{\mu_k,\Sigma_k} \big(x \big) = (\sqrt{2 \pi})^{-p/2} |\Sigma_k|^{-1/2} \exp  \left\{ -(x - \mu_k)^\top \Sigma_k^{-1} (x - \mu_k)/2 \right\}.
$
Here, $|A|$ represents the determinant of an arbitrary square matrix $A.$
Denote
$\theta_0 = (\alpha_k, \mu_k^\top,$\quad$ \vech(\Sigma_k)^\top; k=1,\dots,K)^\top \in \mR^{q}$ as the true parameter of interest, where $q=K(p^2+3p+2)/2$ and $\vech(A)$ represents the half-vec operator with $\vech(A) = (A_{ij} : 1 \leq i \leq j \leq p) \in \mR^{p(p+1)/2}$ for any square matrix $A \in \mR^{p \times p}$.
Then, the PDF of the GMM can be written as
$
f(x|\theta) = \sum_{k=1}^K \alpha_k \phi_{\mu_k,\Sigma_k} \big(x \big).
$
We then have its negative log-likelihood function as
$
\mL(\theta) = N^{-1} \sum_{i=1}^N \ell_i(\theta) =  - N^{-1}\sum_{i=1}^N \log f(X_i|\theta).
$
The maximum likelihood estimator (MLE) can be obtained as $\wh\theta = \argmin_\theta \mL(\theta)$.

To compute $\wh\theta$, the well-known EM algorithm can be employed. A standard EM algorithm can be motivated by the method of complete-data log-likelihood, where the latent class labels $Y_i$s are assumed to be known. In this study, we present a different perspective on the gradient condition. Specifically, we define \(\dot{\mL}(\theta) \in \mathbb{R}^q\) as the derivative of \(\mL(\theta)\) with respect to $\theta$. Recall that $\wh{\theta}$ denotes the MLE. Consequently, we should have a gradient condition: \(\dot{\mL}(\wh{\theta}) = 0\). This leads to a set of estimation equations as
$\wh \alpha_k = N^{-1} \sum_{i=1}^N \wh \pi_{ik}$,
$\wh \mu_k = ( \sum_{i=1}^N \wh \pi_{ik} )^{-1} \sum_{i=1}^N \wh \pi_{ik} X_i$, and
$\wh \Sigma_k = ( \sum_{i=1}^N \wh \pi_{ik} )^{-1} \sum_{i=1}^N \wh \pi_{ik} (X_i - \wh\mu_k)(X_i - \wh\mu_k)^\top$ for $1 \leq k \leq K$.
Here, $\wh \pi_{ik} = f^{-1}(X_i|\wh\theta) \wh \alpha_k \phi_{\wh \mu_k, \wh \Sigma_k}(X_i)$ is the estimator of $\pi_{ik} = f^{-1}(X_i|\theta_0)  \alpha_k \phi_{ \mu_k,  \Sigma_k}(X_i)$. Inspired by the above equations, an iterative algorithm can be obtained as follows.
Let $\wh \theta^{(0)} = (\wh \alpha_{k}^{(0)}, (\wh\mu_{k}^{(0)})^\top,\vech^\top(\wh \Sigma_{k}^{(0)});k=1,\dots,K)^\top$ be an initial estimator. Let $\wh \theta^{(t)} = (\wh \alpha_{k}^{(t)}, (\wh\mu_{k}^{(t)})^\top$, 
$\vech^\top(\wh \Sigma_{k}^{(t)});k=1,\dots,K)^\top$ be the estimator obtained in the $t$th step. Define $\wh \pi^{(t)}_{ik} = \wh \alpha^{(t)}_k$ 
$\phi_{\wh \mu^{(t)}_k, \wh \Sigma^{(t)}_k}(X_i) f^{-1}(X_i|\wh\theta^{(t)})$ as the estimator of $\pi_{ik}$ by using the estimator obtained in the $t$th iteration.
We then obtain the $(t+1)$th iteration estimator $\wh \theta^{(t+1)} = (\wh \alpha_{k}^{(t+1)}, (\wh\mu_{k}^{(t+1)})^\top$, 
$\vech^\top(\wh \Sigma_{k}^{(t+1)});k=1,\dots,K)^\top$ as
\beqr
\label{eq:em1}
\wh \alpha^{(t+1)}_k  &=& N^{-1} \sum_{i=1}^N \wh \pi^{(t)}_{ik},\quad 
\wh \mu^{(t+1)}_k = \Big( \sum_{i=1}^N \wh \pi^{(t)}_{ik} \Big)^{-1} \sum_{i=1}^N \wh \pi^{(t)}_{ik} X_i ,  \\
\label{eq:em2}
\wh \Sigma^{(t+1)}_k &=& \Big( \sum_{i=1}^N \wh \pi^{(t)}_{ik} \Big)^{-1} \sum_{i=1}^N \wh \pi^{(t)}_{ik} \Big(X_i - \wh\mu^{(t+1)}_k\Big)\Big(X_i - \wh\mu^{(t+1)}_k\Big)^\top .
\eeqr
One can then iterate \eqref{eq:em1} and \eqref{eq:em2} until $\wh \theta^{(t+1)}$ converges numerically.
As one can see, this is indeed a standard
EM algorithm \citep{wu1983convergence}. Its numerical convergence has been extensively studied in the literature \citep{chen1995optimal,JMLR:v25:23-1245}.

\subsection{A Na\"ive Network EM} \label{subsec:nnem}

We next extend the classic EM algorithm to DFL.
Define the whole dataset as $\mS = \{1,2,\dots,$
$N\}$, which is distributed on a total of $M$ sites. Collect the indices of the sample allocated to the $m$th site by $\mS_m \subset\mS.$ We then have $\mS = \cup_{m=1}^M \mS_m$ and $\mS_{m_1}\cap \mS_{m_2}=\emptyset$ for any ${m_1}\not ={m_2}$. For convenience, we assume that $|\mS_m| = n  = N/M$ for every $1 \leq m \leq M.$
Next, denote the indices of the sites by $\mathcal{M} = \{1,2,\dots,M\}$. These sites are connected by a communication network $A = (a_{m_1 m_2}) \in \mR^{M \times M}$, where $a_{m_1m_2} = 1$ if site $m_1$ can receive information from $m_2$, and $a_{m_1m_2} = 0$ otherwise. Following tradition, we assume that $a_{mm} = 0$ for $1 \leq m \leq M$.
Define a weighting matrix as $W = (w_{m_1 m_2}) \in \mR^{M \times M}$ with $w_{m_1 m_2} = a_{m_1 m_2} / d_{m_1}$, where $d_{m_1} = \sum_{m_2} a_{m_1 m_2}$ is the in-degree for site $m_1$. Here, we assume $d_m > 0$ for every $1 \leq m \leq M.$ Otherwise, the corresponding site is isolated from the remaining network and thus cannot participate in the algorithm. Then, the global loss function $\mL(\theta)$ can be rewritten as
$
\mL(\theta) = M^{-1} \sum_{m=1}^M \mL_{(m)} (\theta),
$
where $\mL_{(m)} (\theta) = n^{-1} \sum_{i \in \mS_m} \ell_i(\theta)$ is the loss function defined on the $m$th site. We next consider how to optimize $\mL(\theta)$ in a decentralized way.

To this end, the classic EM algorithms in \eqref{eq:em1} and \eqref{eq:em2} need to be revised subsequently. First, we introduce a new notation: $(\wh \alpha_{k}^{(t,m)}, (\wh\mu_{k}^{(t,m)})^\top, \vech^\top(\wh \Sigma_{k}^{(t,m)});$ $k=1, \dots, K)^\top$, which represents the estimator computed in the $t$th iteration on the $m$th site. Under the DFL framework, the $m$th site cannot receive information from the entire network. Instead, it only receives information from its connected neighbors. This leads to a neighborhood averaged estimator $\wt\alpha_{k}^{(t,m)} = \sum\nolimits_{q=1}^M w_{mq} \wh \alpha_{k}^{(t,q)}, \wt\mu_{k}^{(t,m)} =  \sum\nolimits_{q=1}^M w_{mq} \wh \mu_{k}^{(t,q)},$ and $\wt\Sigma_{k}^{(t,m)} = \sum\nolimits_{q=1}^M w_{mq} \wh \Sigma_{k}^{(t,q)}$.
Next, the local estimator at the $(t+1)$th iteration can be computed as:
\beqr
\label{eq:nnem}
\wh \alpha^{(t+1,m)}_{k}  &=& \sum\nolimits_{i\in\mS_m} \wt \pi^{(t,m)}_{ik}/n, \quad \wh \mu^{(t+1,m)}_{k} = n^{-1}\sum\nolimits_{i\in\mS_m} \wt \pi^{(t,m)}_{ik} X_i/ \wh \alpha^{(t+1,m)}_{k},   \\
\wh \Sigma^{(t+1,m)}_{k} &=& n^{-1}\sum\nolimits_{i\in\mS_m} \wt \pi^{(t,m)}_{ik} \Big(X_i - \wt\mu^{(t,m)}_{k}\Big)\Big(X_i - \wt\mu^{(t,m)}_{k}\Big)^\top/\wh \alpha^{(t+1,m)}_{k}. \nonumber
\eeqr
Here, $\wt \pi^{(t,m)}_{ik} = f^{-1}\big(X_i|\wt\theta^{(t,m)}_{k}\big) \wt \alpha^{(t,m)}_{k} \phi_{\wt \mu^{\scalebox{0.4}{$(t, m)$}}_{k}, \wt \Sigma^{\scalebox{0.4}{$(t, m)$}}_{k}}(X_i).$ The algorithm can then be iterated until it converges numerically. For convenience, we refer to this algorithm as the Na\"ive Network Expectation–Maximization (NNEM) algorithm.
The NNEM method is easy to describe and implement. It can also be proven that NNEM achieves consistent estimation under stringent conditions, requiring data to be homogeneously distributed across sites. 
The relevant theory is provided in Web Appendix B.8. 
However, our analysis suggests that the NNEM estimator might be biased with heterogeneous data; see Web Appendix B.8 for a quick understanding.

\section{A Momentum-based Network EM Algorithm} \label{sec:methodology}

\subsection{The MNEM Algorithm} \label{sec:MNEM:alg}

According to Section \ref{subsec:nnem}, the NNEM estimator is statistically inconsistent with heterogeneously distributed data.
We next examine the potential causes of this unsatisfactory performance for a remedy. 
Notably, for most decentralized gradient descent algorithms, the bias introduced by heterogeneous data can be effectively mitigated by selecting a sufficiently small learning rate \citep{wu2023network}. 
This is because the information can be passed from one site to another with a long distance on the network with a sufficiently small learning rate. 
However, in the NNEM algorithm outlined in \eqref{eq:nnem}, both the E-step and M-step calculations utilize closed-form expressions without a learning rate type of tuning parameter involved. The consequence is that we lose the opportunity to control the decay rate for information exchange across different sites in the network, which in turn leads to seriously biased estimation results. 

Therefore, we are inspired to develop a momentum-based network EM (MNEM) algorithm as follows. Unlike the NNEM algorithm, the MNEM algorithm
incorporates a momentum effect. The momentum effect allows one to update a local estimator via estimators from not only the current step but also the previous step.
With the help of the momentum effect, the information from less closely connected sites can be exchanged in a more efficient way. This often leads to algorithms with faster convergence.
Specifically, let $\wh \theta^{(t,m)}_{\mnem}=\big(  \wh \alpha_{\mnem,k}^{(t,m)}$, $\big(\wh\mu_{\mnem,k}^{(t,m)}\big)^\top, \vech^\top\big(\wh\Sigma_{\mnem,k}^{(t,m)}\big);k=1,\dots,K\big)^\top$
be the MNEM estimator obtained after the $t$th iteration on the $m$th site. We then update $\wh \theta^{(t,m)}_{\mnem}$ to be $\wh \theta_{\mnem}^{(t+1,m)}$ by the following two steps.

{\sc Step 1.}
Similar to the NNEM algorithm, each site updates the estimators utilizing information from its neighbors. However, we slightly modify the first step of the NNEM algorithm following the ideas of \cite{gu2008distributed} to reduce the bias of the neighborhood-averaged estimators. Specifically, each site calculates $\wh \beta^{(t,m)}_{\mnem,k} = \wh \mu_{\mnem,k}^{(t,m)} \wh \alpha_{\mnem,k}^{(t,m)}$ and $\wh \gamma^{(t,m)}_{\mnem,k} = \wh \Sigma_{\mnem,k}^{(t,m)} \wh \alpha_{\mnem,k}^{(t,m)}$. Next, $\wh \beta_{\mnem,k}^{(t,m)}$ and $\wh \gamma_{\mnem,k}^{(t,m)}$, instead of $\wh \mu_{\mnem,k}^{(t,m)}$ and $\wh \Sigma_{\mnem,k}^{(t,m)}$, are passed to neighbors so that neighborhood-averaged estimators can be computed as follows:
\beq
\label{eq:mnem:avg}
\wt\alpha_{\mnem,k}^{(t,m)} = \sum_{q=1}^M w_{mq} \wh \alpha_{\mnem,k}^{(t,q)}, \
\wt\beta_{\mnem,k}^{(t,m)} = \sum_{q=1}^M w_{mq} \wh \beta_{\mnem,k}^{(t,q)}, \ \wt\gamma_{\mnem,k}^{(t,m)} = \sum_{q=1}^M w_{mq} \wh \gamma_{\mnem,k}^{(t,q)}. \nonumber
\eeq

{\sc Step 2.} Once the neighborhood-averaged estimators (i.e., $\wt\alpha_{\mnem,k}^{(t,m)}, \wt\beta_{\mnem,k}^{(t,m)}, \wt\gamma_{\mnem,k}^{(t,m)}$) are obtained by the $m$th site, they are then further updated by a momentum-based algorithm, which incorporates the information from the local sites as follows.
\beqr
\wh \alpha^{(t+1,m)}_{\mnem,k}  &=& \frac{\eta}{n}  \sum_{i\in\mS_m} \wt \pi^{(t,m)}_{ik} + (1-\eta) \wt \alpha^{(t,m)}_{\mnem,k}, \  \wh \beta^{(t+1,m)}_{\mnem,k} =  \frac{\eta}{n} \sum_{i\in\mS_m} \wt \pi^{(t,m)}_{ik} X_i  +  (1-\eta) \wt \beta^{(t,m)}_{\mnem,k}, \nonumber\\ 
\wh \gamma^{(t+1,m)}_{\mnem,k} &=&  \frac{\eta}{n} \sum_{i\in\mS_m} \wt \pi^{(t,m)}_{ik} \Big(X_i - \wt\mu^{(t,m)}_{\mnem,k} \Big)\Big(X_i - \wt\mu^{(t,m)}_{\mnem,k}\Big)^\top  + (1-\eta) \wt \gamma^{(t,m)}_{\mnem,k}. \label{eq:MNEM:update}
\eeqr
Here $\wt \pi^{(t,m)}_{ik} = f^{-1}(X_i|\wt\theta^{(t,m)}_{\mnem}) \wt \alpha^{(t,m)}_{\mnem,k} \phi_{\wt \mu^{\scalebox{0.4}{$(t, m)$}}_{\mnem,k}, \wt \Sigma^{\scalebox{0.4}{$(t, m)$}}_{\mnem,k}}(X_i)$ with $\wt\mu^{(t,m)}_{\mnem,k} = \wt\beta^{(t,m)}_{\mnem,k}/  \wt\alpha^{(t,m)}_{\mnem,k}; \wt\Sigma^{(t,m)}_{\mnem,k}$ 
$= \wt\gamma^{(t,m)}_{\mnem,k}/  \wt\alpha^{(t,m)}_{\mnem,k}$ is an estimator for $\pi_{ik}$, and
$\eta > 0$ is a tuning parameter. Comparing \eqref{eq:MNEM:update} with \eqref{eq:nnem}, one key difference is the introduction of this momentum effect (i.e., neighborhood-averaged estimators). The momentum effect is
controlled by the tuning parameter $\eta.$ For convenience, we refer to $\eta$ as a momentum parameter. With $\eta = 1$, the above algorithm \eqref{eq:MNEM:update} reduces to the one without the momentum effect. In contrast, if a smaller $\eta$ is used, a stronger momentum effect is allowed. By including this momentum effect appropriately, the parameter update of MNEM becomes less reliant on the local data. Therefore, the MNEM algorithm is more likely to borrow information from those sites, which are not closely connected. This therefore alleviates the bias due to heterogeneity.

\subsection{Theoretical Properties of MNEM}\label{sec:theorem:MNEM}

We first introduce the notations and metrics. Denote the MNEM estimator obtained at iteration $t$ on the
$m$th site as $ \wh \psi_{\mnem}^{(t,m)} = \big( \wh \alpha_{\mnem,k}^{(t,m)},(\wh \beta_{\mnem,k}^{(t,m)})^\top,\vech^\top(\wh \gamma_{\mnem,k}^{(t,m)});k=1,\dots,K \big)^\top \in \mR^q$. Each iteration in \eqref{eq:MNEM:update} can then be simplified as
$
\wh \psi_{\mnem}^{(t+1,m)} = (1-\eta) \wt\psi_{\mnem}^{(t,m)} + \eta \mathcal{F}_{(m)}( \wt\psi_{\mnem}^{(t,m)} ),
$
where $\mathcal{F}_{(m)}(\cdot)$ is a mapping function, and its analytical form is provided in Web Appendix A.2. 
Let $\wh \psi\in\mR^q$ and $\psi_0\in\mR^q$ be the corresponding whole-sample estimator and the true parameter, respectively. Next, define $\widehat{ \operatorname{SE}}^2 (W) = M^{-1}\| W^\top \textbf{1}_M - \textbf{1}_M \|^2$
with $\textbf{1}_M=(1,\dots,1)^\top \in\mR^M.$ As noted by \cite{wu2023network}, this is an interesting measure for the balance of network structures.
Intuitively, a smaller $\SEW$ value implies a more balanced network.
In the most ideal case, where $W$ is doubly stochastic in the sense that $\textbf{1}_M^\top W = \textbf{1}_M^\top$ \citep{yuan2016convergence}, we have \(\SEW = 0\).
Further, define $\sigma_w^2 = \| W^\top (I_M - M^{-1} \textbf{1}_M \textbf{1}_M^\top ) W  \|$. 
Notably, this metric is closely related to the numerical convergence rate of a sequence $W^t$ to its limit as $t \to \infty.$
To illustrate, we temporarily assume that $W$ is a doubly stochastic matrix.
It follows that $\|W^{t+1} - W^{\infty}\| \leq \sigma_w^t \|W - W^\infty\|$; see Web Appendix B.6 for details.
Thus, a smaller $\sigma_w$ suggests that $W$ can converge to $W^{\infty}$ at a faster speed. We define $\widehat{ \operatorname{SE}}^2(\mF) = M^{-1}\sum_{m=1}^M \| \mF_{(m)}(\wh\psi) -  \mF(\wh\psi) \|^2$,
where $\mF(\wh\psi) = M^{-1} \sum_{m=1}\mF_{(m)}(\wh\psi)$. The
$\widehat{ \operatorname{SE}}^2(\mF)$ gauges the data heterogeneity. When the data are homogeneous, we have $\widehat{ \operatorname{SE}}^2(\mF) = O_p(1/n)$. In contrast, a larger $\widehat{ \operatorname{SE}}(\mF)$ value indicates a more heterogeneous data distribution. 

We write $\wh \psi^{\;*(t)}_{\mnem} = \big\{ ( \wh \psi^{(t,1)}_{\mnem} )^\top,\dots,( \wh \psi^{(t,M)}_{\mnem} )^\top \big\}^\top \in \mR^{Mq}$ and $\wh \psi^* = I^* \wh\psi \in \mR^{Mq}$ as the stacked MNEM estimator and whole sample estimator, respectively.
Let
$
\wh\delta_{0}^{(m)} =  \| \wh \psi_{\mnem}^{(0,m)} - \wh\psi \|
$
represent the estimation error due to the initial estimator. For simplicity, assume that all sites start from the same initial value $\wh\delta_{0}^{(m)} \equiv  \wh \delta_0$.
We then have the following theorem.
\begin{theorem}
\label{thm:mnem}
Assume conditions (C1), (C2.a), and (C3)–(C7) in Web Appendices B.1 and B.4 hold. 
Further, assume that $\eta + \SEW$ is sufficiently small and the initial value $\wh \psi^{\;*(0)}_{\mnem}$ is sufficiently close to $\wh\psi^*$ in the sense that $\eta + \SEW < \epsilon$ and $\| \wh \delta_0 \| \leq \epsilon$ for some sufficiently small but fixed positive constant $\epsilon$. Then, $M^{-1/2} \|\widehat{\psi}^{\;*(t)}_{\mnem} - \wh\psi^*\|$ is upper bounded by
\beq
\label{eq:mnem:non-asymp}
C_1 \left[ \left\{ 1- \frac{\eta}{2} \left(1 - \bar{L}_{\max}\right)\right\}^{t-1} \wh\delta_0 + \frac{\eta  + \SEW }{\big(1 - \bar{L}_{\max}\big)(1 - \rho_w)} \SEF +  \frac{\big\{\eta  + \SEW\big\}^2 }{\big(1 - \bar{L}_{\max}\big)^2(1 - \rho_w)^2} \right]
\eeq
with probability at least $1 - O\big(M \exp(-c_1 n^r)\big)$ for some fixed $0 < r \leq 1$ and some fixed  constants $c_1,C_1$. Here $c_1, C_1$ are independent of $(\bar{L}_{\max},n,M).$
\end{theorem}

From Theorem \ref{thm:mnem}, the estimation error of MNEM can be decomposed into two parts. The first part in \eqref{eq:mnem:non-asymp} is the numerical error due to $\wh \delta_0$, which converges toward $0$ exponentially. This numerical error is mainly regularized by the momentum parameter $\eta$ and the separability of the Gaussian components through the constant $\bar{L}_{\max}$. 
Specifically, a better separability condition implies a smaller $\bar{L}_{\max}$. Moreover, a smaller $\bar{L}_{\max}$ and a larger $\eta$ lead to faster numerical convergence. 
Notably, we cannot set $\eta = 0$. 
When $\eta = 0$, this part becomes a constant and thus does not converge to $0$. 
This finding highlights the importance of the momentum effect in the MNEM algorithm.
The second part in \eqref{eq:mnem:non-asymp} is the statistical error controlled by the momentum effect \(\eta\), the network structure $\SEW$, the data heterogeneity effect \(\SEF\), and the separability conditions $\bar{L}_{\max}$. 
Thus, the statistical efficiency of the MNEM estimator can be improved by (i) a smaller but positive momentum parameter $\eta$; (ii) a more balanced network structure $W$; (iii) better separability conditions; and (iv) a more homogeneous data distribution.
Further, the MNEM estimator can be asymptotically as efficient as the whole sample estimator, as long as $\eta + \SEW = o_p(N^{-1/2})$.
This condition can be satisfied for sufficiently balanced networks and small momentum parameter $\eta$.

\subsection{Semi-supervised MNEM Algorithm}\label{sec:semi-MNEM:alg}

The MNEM method developed in the previous subsection can effectively handle heterogeneous data. However, it requires the operator $\Mean \big\{ \dot{\mF}(\psi_0) \big\}$ to be a contraction operator (i.e., Condition (C3) holds), which may not hold if different Gaussian components are poorly separated; see Web Appendix B.7 for a more detailed explanation. 
To address this, we further develop below a semi-supervised MNEM algorithm by incorporating information from partially labeled data.
We refer to this new approach as the semi-MNEM method.

Specifically, the semi-MNEM is similar to the MNEM. The only difference is that the information provided by partially annotated labels is used by the semi-MNEM. Specifically, assume the existence of a subset of subjects $\mS^*\subseteq \mS=\{1,2,\cdots,N\}$ where both the feature $X_i$ and the label $Y_i$ are observed.
This leads to the following 
negative log-likelihood function in the semi-supervised learning setting:
\beq
\label{eq:semi:likelihood}
\mL_{\semi}(\theta) = - \frac{1}{N}  \sum_{i\notin\mS^*} \log f(X_i|\theta) - \frac{1}{N}
\sum_{i\in\mS^*} \sum_{k=1}^{K}  \mathbb{I}(Y_i = k) \Big\{ 
\log \phi_{\mu_k,\Sigma_k}(X_i) + \log \alpha_k\Big\}. \nonumber
\eeq 
Let $\mS_m^* \subset \mS^*$ be the index set of all labeled observations on the $m$th site. For simplicity, we assume that   $|\mS_m^*| = n^*$ for every $1 \leq m \leq M.$ Define
$
\wh \theta_{\snem}^{(t,m)} = \big( \wh \alpha_{\snem,k}^{(t,m)}, (\wh \mu_{\snem,k}^{(t,m)})^\top,\vech^\top$\quad$\big(\wh \Sigma_{\snem,k}^{(t,m)}\big); k = 1,\dots,K  \big)^\top \in \mR^q 
$
as the semi-MNEM estimator on the $m$th site at the \(t\)th iteration.
We then update $\wh \theta_{\snem}^{(t+1,m)}$ according to:  
$
\wh \alpha^{(t+1,m)}_{\snem,k} = \eta \big( \sum_{i \in \mS_m\backslash \mS_m^*}  \wt \pi^{(t,m)}_{ik} + \sum_{i \in \mS_m^*} a_{ik} \big)/n + (1-\eta) \wt \alpha^{(t,m)}_{\snem,k},$ 
$ 
\wh \beta^{(t+1,m)}_{\snem,k} = \eta \big( \sum_{i \in \mS_m\backslash \mS_m^*}  \wt \pi^{(t,m)}_{ik} X_i + \sum_{i\in\mS^{*}_{m}} a_{ik} X_i \big)/n  +  (1-\eta) \wt \beta^{(t,m)}_{\snem,k},$ and
\beqr
\label{eq:SNEM:update}
\wh \gamma^{(t+1,m)}_{\snem,k} &=& (1-\eta) \wt \gamma^{(t,m)}_{\snem,k} + \frac{\eta}{n} \bigg\{\sum_{i \in \mS_m\backslash \mS_m^*} \wt \pi^{(t,m)}_{ik} \Big(X_i - \wt\mu^{(t,m)}_{\snem,k} \Big)\Big(X_i - \wt\mu^{(t,m)}_{\snem,k}\Big)^\top \nonumber \\
&+& \sum_{i\in\mS_m^*} a_{ik} \Big(X_i - \wt\mu^{(t,m)}_{\snem,k} \Big)\Big(X_i - \wt\mu^{(t,m)}_{\snem,k}\Big)^\top \bigg\}.
\eeqr
where $
a_{ik} = \mathbb{I}(Y_i = k)$ is the information provided by partially labeled data. Moreover, $\wt\beta_{\snem,k}^{(t,m)},\wt\gamma_{\snem,k}^{(t,m)}$, and $\wt\alpha_{\snem,k}^{(t,m)}$ are the neighborhood-averaged estimators.
Compared with the updating formulas of MNEM in \eqref{eq:MNEM:update},
those of semi-MNEM in \eqref{eq:SNEM:update} integrate information from both features \(X_i\)
and \(a_{ik}\), where the latter contains valuable information from partially annotated labels. This integration leads to a contraction operator with a reduced $\ell_2$-norm,
which thereby improves semi-MNEM's numerical convergence
over MNEM. This is particularly helpful if some mixture components are not well separated. Further details of semi-MNEM are relegated to Algorithm C.1 in Web Appendix C.1.

We next establish the theoretical properties of semi-MNEM. To this end, define $ \wh \psi_{\snem}^{(t,m)} = \big( \wh \alpha_{\snem,k}^{(t,m)}, (\wh \beta_{\snem,k}^{(t,m)})^\top, \vech^\top(\wh\gamma_{\snem,k}^{(t,m)});k=1,\dots,K \big)^\top \in \mR^q$. Given the neighborhood averaged estimator $\wt\psi_{\snem}^{(t,m)}$, each iteration in \eqref{eq:SNEM:update} can be represented as
$
\wh \psi_{\snem}^{(t+1,m)} = (1-\eta) \wt\psi_{\snem}^{(t,m)} + \eta \msF_{(m)}(\wt\psi_{\snem}^{(t,m)})
$
for some mapping function \(\msF_{(m)}( \cdot) \), whose analytical form is given in Web Appendix A.2. 
Denote $\bar{L}^{\semi}_{\max} = \big\|\Mean\big\{ \dot{\msF}(\psi_0) \big\}\big\|$ with $\dot{\msF}(\psi_0) = M^{-1} \sum_{m=1}^M \dot{\msF}_{(m)}(\psi_0)$ and $\wh \psi_{\snem}^{\;*(t)} = \big\{(\wh \psi_{\snem}^{(t,1)})^\top ,$\quad$\dots,  (\wh \psi_{\snem}^{(t,M)})^\top \big\}^\top \in \mR^{Mq}$ as the stacked semi-MNEM estimate obtained in the $t$th iteration. Write $\wh \psi_{\semi}$ and $\wh \psi^*_{\semi} = I^* \wh\psi_{\semi} \in \mR^{Mq}$ as the whole sample estimator and its stacked counterpart, respectively. Let
$
\wh\delta_{0}^{\;\dagger(m)} =  \| \wh \psi_{\mnem}^{(0,m)} - \wh\psi \|
$
and assume that all sites start from the same initial value $\wh \delta_{0}^{\;\dagger(m)} \equiv \wh \delta_{0}^{\;\dagger}$.
We summarize our results in the following theorem.
\begin{theorem}
\label{col:snem}
Assume conditions (C1), (C2.b), (C4), and (C5) -- (C7) in Web Appendices B.1 and B.4 hold. Further, assume $\eta + \SEW$ is sufficiently small and the initial value $\wh \psi^{\;*(0)}_{\snem}$ is sufficiently close to $\wh\psi^*_{\semi}$ in the sense that $\eta + \SEW < \epsilon$ and $\| \wh \delta_0^\dagger \| \leq \epsilon$ for some sufficiently small but fixed positive constant $\epsilon$. We then have
$
\bar{L}^{\semi}_{\max} =  \left( 1 - n^*/n\right) \bar{L}_{\max}
$ and $\bar{L}^{\semi}_{\max} < 1$ for a sufficiently large $n^*$. In addition, we have $M^{-1/2} \|\widehat{\psi}^{*(t)}_{\snem} - \wh\psi^*_{\semi}\|$ upper bounded by
\beq
\label{eq:snem:non-asymp}
C_2 \left[ \left\{ 1- \frac{\eta}{2} \left(1 - \bar{L}_{\max}^{\semi}\right)\right\}^{t-1} \wh\delta_0^{\dagger} + \frac{\eta  + \SEW }{\big(1 - \bar{L}_{\max}^{\semi}\big)(1 - \rho_w)} \SEsF +  \frac{\big\{\eta  + \SEW\big\}^2 }{\big(1 - \bar{L}_{\max}^{\semi}\big)^2(1 - \rho_w)^2} \right]
\eeq
with probability at least $1 - O\big(M \exp(-c_2 n^r)\big)$ for some fixed $0 < r \leq 1$  and some constants $c_2,C_2>0$. Here $c_2, C_2$ are independent of $(\bar{L}^{\semi}_{\max},n,M).$
\end{theorem}
\noindent
Comparing equation \eqref{eq:snem:non-asymp} against \eqref{eq:mnem:non-asymp}, we observe that the key difference is that \(\bar{L}^{\semi}_{\max} = (1 - n^*/n) \bar{L}_{\max} < \bar{L}_{\max}\) due to partially labeled instances. This leads to a much improved numerical convergence rate even if the mixture components are poorly separated. Similar interesting findings were also obtained by \cite{JMLR:v25:23-1245} but under a different GMM setup.

\section{Numerical Studies}
\label{sec:Numerical}
\subsection{Simulation Setup} \label{sec:simu}

To evaluate the finite sample performance, we present several simulation studies. Specifically, we set the feature dimension as $p=6$, the number of categories as $K=3$, the sample size as $N = 30{,}000$, and the number of sites as $M = 20$. 
This results in a sample size of $n=1{,}500$ for each site.
We next generate $Y_i$ according to $\mathbb{P}(Y_i=k)=\alpha_k$ for $1\leq k \leq 3$ with $\alpha_1 =0.5$, $\alpha_2=0.3$, and $\alpha_3=0.2$.
Given $Y_i=k$, $X_i$ is then generated from the multivariate Gaussian distribution $N_p(\mu_k,\Sigma_k)$. Here, \(\Sigma_k = (\sigma_{k,ij})\) with \(\sigma_{k,ij} = \rho_k^{|i-j|}\), where \(\rho_1 = 0.5\), \(\rho_2 = 0.1\), and \(\rho_3 = -0.1\). The mean vector $\mu_1$ is simulated from $N_p(0, I_p)$. We set \(\mu_k = \mu_{k-1} + \Delta^*\mathbf{1}_p\) for \(k \in \{2, 3\}\). 
Note that $\Delta^*$ controls the degree of separation, a larger $\Delta^*$ indicates better separated Gaussian mixtures. 
Here we set $\Delta^*\in\{1,2,4\}$ and $\eta \in \{0.01,0.02,0.05\}$. 
To avoid numerical convergence failure, practical guidance for diagnosing such issues is provided in Web Appendix C.8. 
Additionally, in practice, the choice of the momentum parameter $\eta$ is also critical. 
To address this, the detailed practical guidance is provided in Web Appendix C.5.
Moreover, set \(r = n^{*}/n \in \{0, 0.05, 0.10, 0.50, 1\} \) as the annotation ratio, i.e., the fraction of labeled samples. The cases with \(r = 0 \) correspond to the fully unsupervised methods (i.e., the NNEM and MNEM methods), whereas the cases with \(r > 0 \) correspond to the semi-MNEM method.
Here we fix the fractions of labeled data to be the same for each site to be consistent with our theory. 
The situation with heterogeneous fractions of labeled data for different sites is also of great practical importance and therefore to be numerically studied in Web Appendix C.4. 
We find that the results remain encouraging. 

\begin{figure}[h]
\centering
\includegraphics[width=6in]{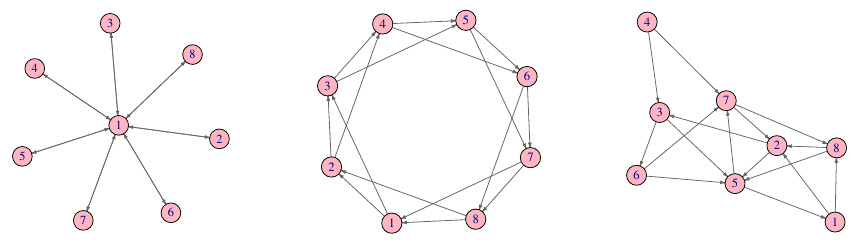}
\caption{Graphical illustration of three important network structures. Left panel: the star network; middle panel: the circle-type network; and right panel: the fixed-degree network.}
\label{fig:1}
\end{figure}

Once the data are simulated, they are distributed to different sites. We consider two different data distribution mechanisms. The first is a homogeneous process, where all observations are distributed to different sites in a completely random way. 
The second is the heterogeneous process, where all observations are sorted according to their response values.
These sorted observations are then distributed sequentially to different sites. 
In addition, three different network structures are considered. They are, namely, the star network structure, the fixed-degree network structure, and the circle-type network structure \citep{wu2023network}.
A graphical illustration of those network structures is given in Figure \ref{fig:1}.
More details of these network structures are presented in Web Appendix C.3.
It can be verified that
the circle-type network structure is the most balanced one with $\SEW=0$, whereas the star network structure is the most unbalanced one with $\SEW=(M-2)/{(M-1)}^{1/2}$; see \cite{wu2023network}. We set various momentum parameters for MNEM and semi-MNEM. 


\begin{figure}[!ht]
\centering
\includegraphics[width=6in]{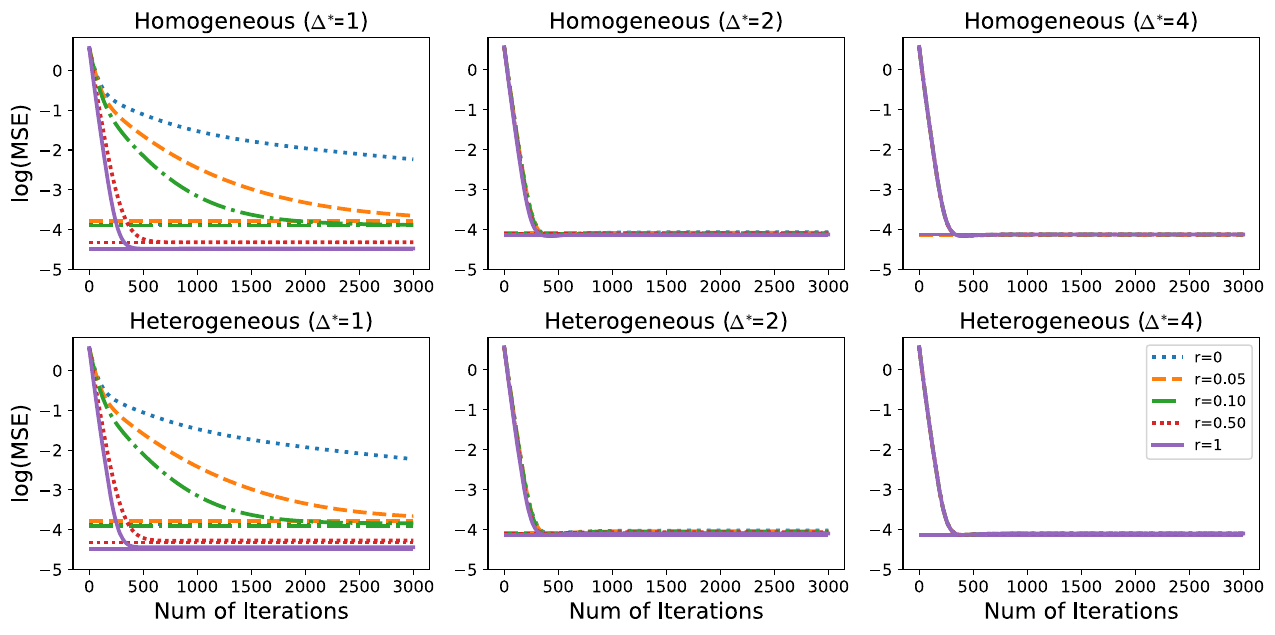}
\caption{The log(MSE) values of MNEM and EM ($r=0$), as well as semi-MNEM and semi-EM ($r>0$) for the circle-type network structure. Each $r$ value corresponds to two lines, where the curved line represents the proposed MNEM or semi-MNEM and the horizontal line represents the whole sample estimator EM or semi-EM. The upper and lower panels represent the homogeneous and heterogeneous data generation processes, respectively. The left, middle, and right panels represent $\Delta^*=1,2,4$, respectively.}
\label{fig2}
\end{figure}

For each simulation study, the experiments are replicated for a total of $R = 100$ times for each parameter setup. Let $\widehat{\theta}_s^{(t,m)}$ be one particular estimator (e.g., the MNEM estimator) obtained on the $m$th site in the $s$th replicate on the $t$th iteration. We then define the mean squared error (MSE) as \(\operatorname{MSE} = (\operatorname{MR})^{-1}\sum_{s=1}^{R}\sum_{m=1}^{M}\|\widehat{\theta}_s^{(t,m)} - \theta_0\|^2\), which is then log-transformed and plotted in Figure \ref{fig2}. For comparison purposes, the local unsupervised estimator $\wh \theta_{\loc}$ (i.e., ``Local''), the unsupervised whole sample estimator $\wh \theta$ (i.e., ``EM''), and the semi-supervised whole sample estimator $\wh \theta_{\semi}$ (i.e., ``semi-EM'') are also evaluated and reported in Figure \ref{fig2}. The simulation results of different settings are qualitatively similar. To save space, we report only the case with a circle-type network and $\eta = 0.01.$
More detailed simulation results are given in Web Appendix C.4.

By Figure \ref{fig2}, we obtain the following interesting findings. First, when $\Delta^*=2$ and $\Delta^*=4$, both MNEM and semi-MNEM converge to the whole sample estimator, even if the data generation process across different sites is heterogeneous. This result is consistent with our theoretical claims in Theorems \ref{thm:mnem} and \ref{col:snem}. Second, when $\Delta^* = 1$, the MNEM struggles to converge numerically. Even after the $3{,}000$th iteration, the algorithm has yet to converge, with a log(MSE) of approximately $-2.16$. This convergence issue can be resolved by the semi-MNEM method with a small annotation ratio. For example, when the annotation ratio is $0.1$, the algorithm converges by the $3{,}000$th iteration, yielding a log(MSE) of approximately $-3.82.$ Finally, both the convergence speed and the statistical efficiency of semi-MNEM improve as the annotation ratio increases. For example, when $\Delta^* = 1$ and the data are heterogeneously distributed, the log(MSE) values at the 3,000th iteration for $r = 0, 0.05, 0.1,$ and $0.5$ are given by $-2.16, -3.77, -3.82,$ and $-4.20$, respectively. These findings corroborate our theoretical results in Theorem \ref{col:snem} very well.

\subsection{Comparison with Competing Methods}\label{sec:competing}

\begin{figure}[!ht]
\centering
\includegraphics[width=6in]{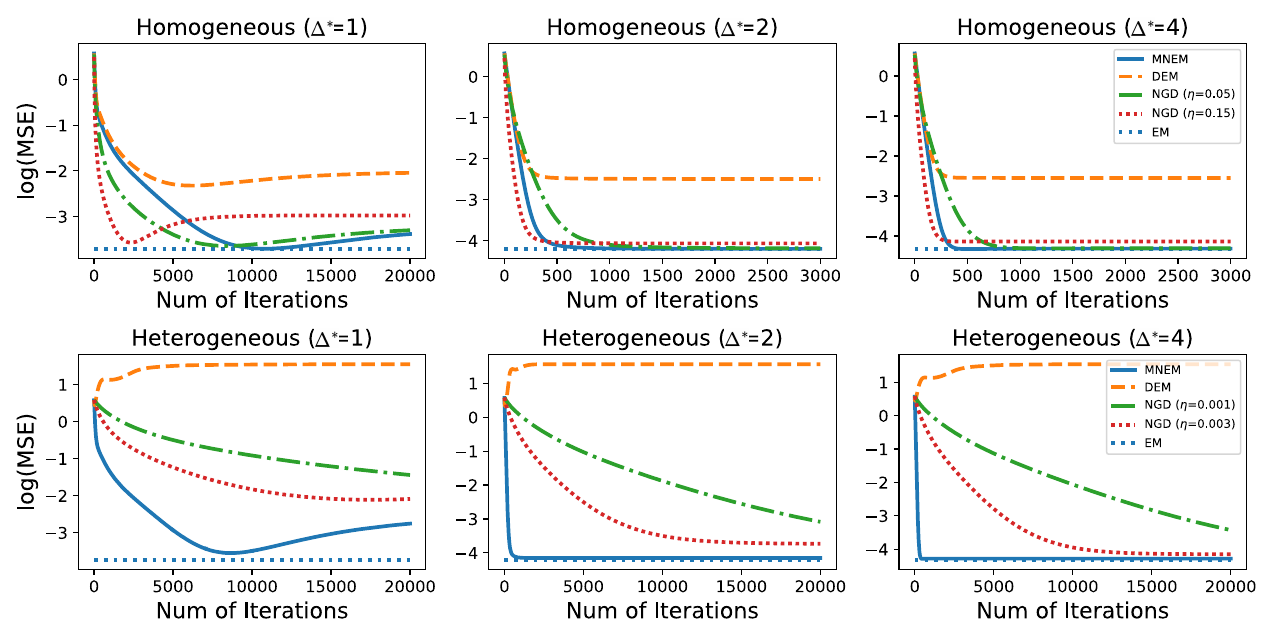}
\caption{The log(MSE) values of MNEM, DEM, and NGD for different distribution patterns and separations of Gaussian components. The blue dotted lines represent the whole sample EM estimators. Here, we fix the network structure to be a circle-type structure and $\eta=0.01$ for the MNEM and DEM. }
\label{fig:competing:unsupervised}
\end{figure}

\begin{figure}[!ht]
\centering
\includegraphics[width=6in]{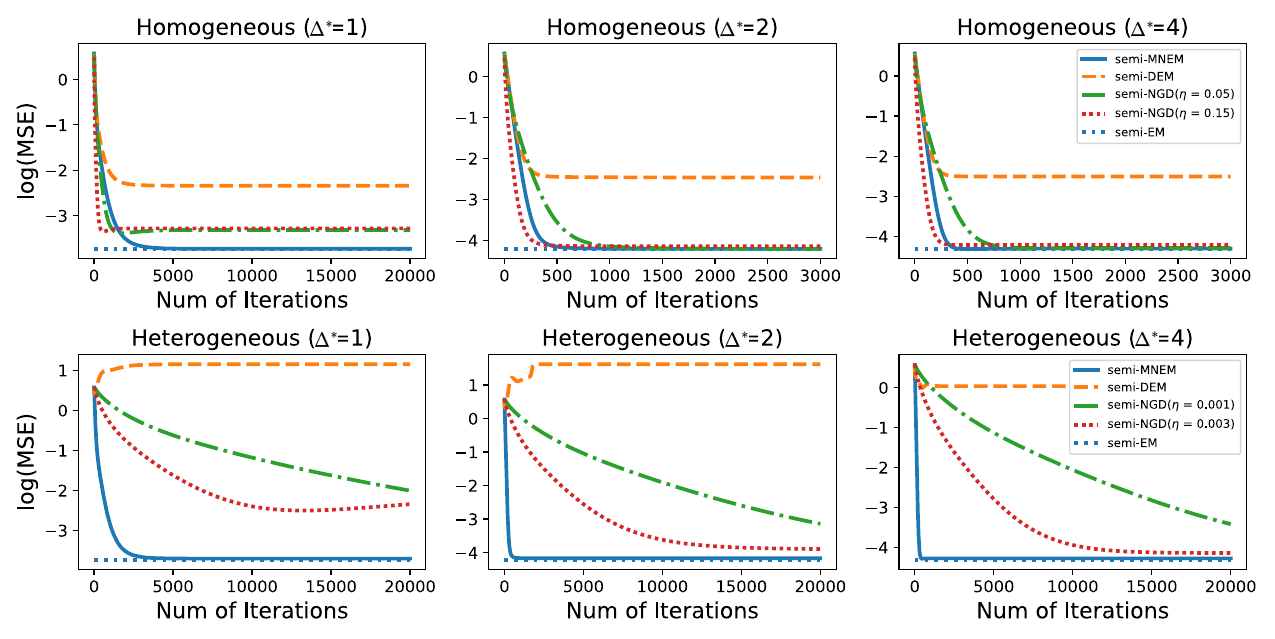}
\caption{The log(MSE) values of the semi-MNEM, semi-DEM, and semi-NGD methods for different distribution patterns and separations of Gaussian components. The blue dotted lines represent the whole sample semi-EM estimators. Here, we fix the network structure to be a circle-type structure, the annotation ratio is $r=0.1$, and $\eta=0.01$ for the MNEM and DEM.}
\label{fig:competing:semi}
\end{figure}

We compare the proposed estimators with two competing methods.
The first method is the distributed EM algorithm over sensor networks (DEM) of \cite{gu2008distributed}.
The second method is a classic decentralized network gradient descent (NGD) method of \cite{yuan2016convergence} and \cite{wu2023network}.
Since both methods can be extended to partially labeled scenarios, we also compare how these two methods perform when the data are partially labeled, denoting the estimators as semi-DEM and semi-NGD. 
The simulation model used here is the same as that in Section \ref{sec:simu}. Specifically, we fix the annotation ratio $r = 0.1$ and the network structure to be a circle-type network. Different \(\eta\) values are carefully selected for (semi-)NGD for their best numerical performance. Here set \(\eta = 0.05,0.15\) for the homogeneous process and \(\eta = 0.001,0.003\) for the heterogeneous process.
For the other methods, we retain \(\eta = 0.01\). The log(MSE) values of MNEM, DEM, and NGD are summarized in Figure \ref{fig:competing:unsupervised}. The log(MSE) values of the semi-MNEM, semi-DEM, and semi-NGD methods are given in Figure \ref{fig:competing:semi}.

We obtain several interesting findings from Figure \ref{fig:competing:unsupervised}. 
First, for the homogeneous case, all methods converge when $\Delta^*$ is relatively large. 
The convergence speed of MNEM is comparable to that of NGD at \(\eta = 0.15\) and clearly better than those of the other two methods. Additionally, DEM yields the highest log(MSE) value, whereas the other methods perform similar to the whole-sample EM estimator. 
When $\Delta^*=1$, only the DEM and NGD methods achieve convergence at $T$= 20,000 steps, yet their log(MSE) values are significantly larger than those of the whole sample estimator. For the heterogeneous case, MNEM outperforms all the other methods when $\Delta^*$ is relatively large. It eventually achieves the same estimation accuracy as the whole sample estimator. However, when $\Delta^*=1$, none of the methods converges by $T$ = 20,000 steps. The log(MSE) values of all methods remain larger than those of the whole-sample estimator.
Lastly, we find that the MSEs go down and then up as the iteration increases. The key reason is that the MNEM algorithm mainly improves parameter estimation accuracy in the early stage (lower MSE), yet faces growing overfitting effects in the later stage (higher MSE). Similar phenomenon has been well documented for many greedy-type machine learning algorithms; see for example the boosting algorithm in \cite{zhang2005boosting}, the gradient descent algorithm in \cite{wu2025benefits}.

From Figure \ref{fig:competing:semi}, we observe qualitatively similar findings with both $\Delta^*=2$ and $\Delta^*=4$. However, with $\Delta^*=1$, we find that the performance of semi-supervised methods is better than that of their unsupervised counterparts. This is mainly due to the utilization of partially labeled data. 
Comparatively speaking, semi-MNEM achieves the smallest log(MSE) value, which is comparable to that of the whole sample estimator. Moreover, the semi-MNEM also has the fastest convergence speed. For example, with $\Delta^* = 2$ and a heterogeneous distribution, the log(MSE) values at the $5,000$th iteration for the semi-DEM, semi-NGD ($\eta = 0.05$), semi-NGD ($\eta = 0.15$), and semi-MNEM are given by $1.20, -1.43, -2.87$ and $-4.10$, respectively.

\subsection{The COVID-19 Dataset}

Finally, we demonstrate our method on the COVID-19 dataset from \cite{chowdhury2020can} and \cite{Tawsifur2021}. 
As one of the largest publicly available COVID-19-positive databases, it contains 21,164 CXR images (299$\times$299 pixels) with four classes: \textit{normal} (10,192 images), \textit{COVID} (3,615 images), \textit{lung opacity} (6,012 images), and \textit{viral pneumonia} (1,345 images). 
This COVID-19 dataset is constructed by integrating samples from a total of eight different public data sources. 
Next, we use the pretrained deep AUC maximization (DAM) model of \cite{yuan2021large} for feature extraction, which achieved the best performance in the CheXpert competition \citep{irvin2019chexpert}.
We adapt this model by removing its fully connected top layers.
This leads to a feature vector of 1,024 dimensions.
Principal component analysis is subsequently applied to further reduce the dimensionality.
By setting the cumulative variance contribution to 90\%, the resulting feature dimension is reduced to $p = 95$.

\begin{figure}[!h]
\centering  
\includegraphics[width=4in]{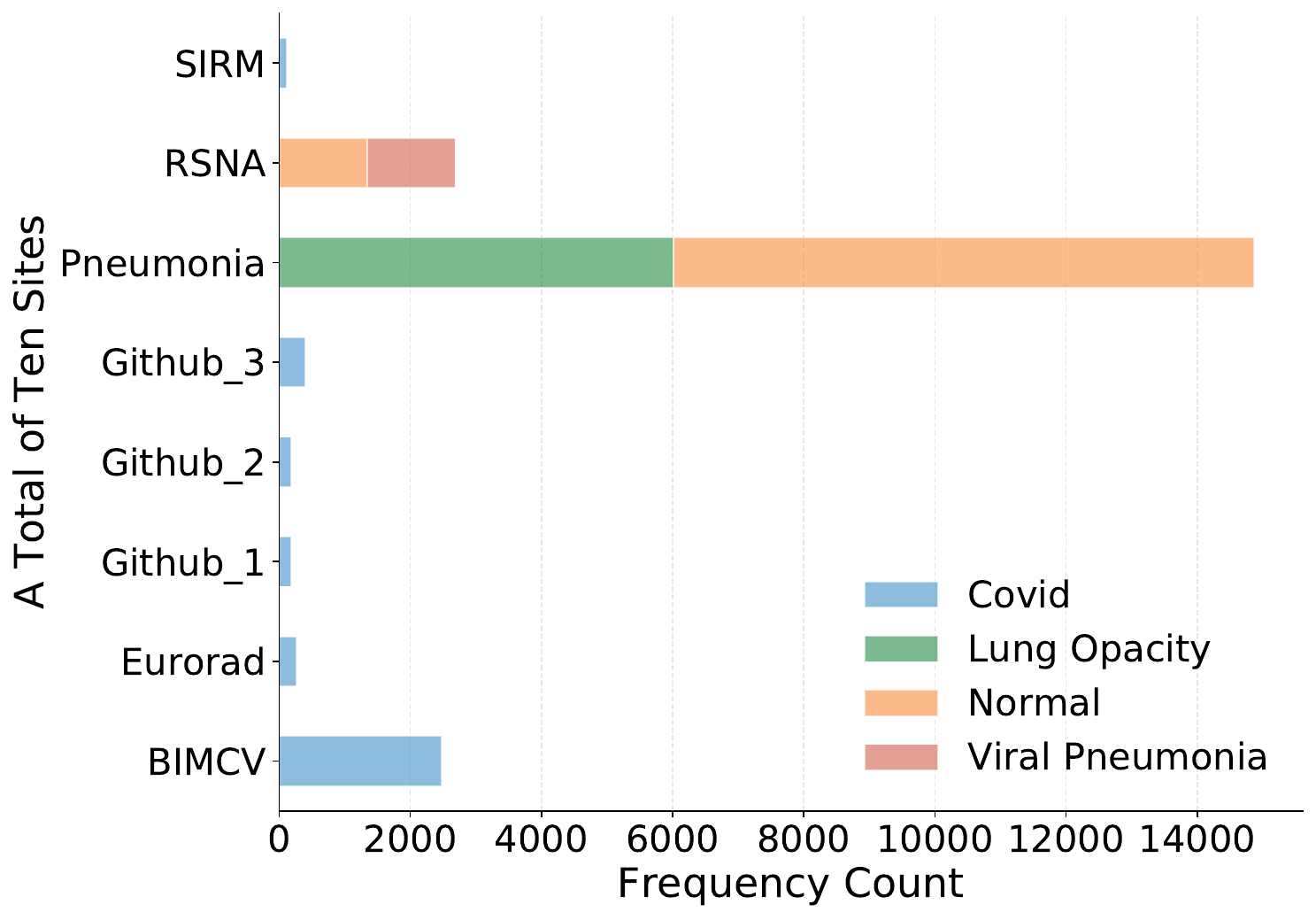}
\caption{The label distributions of the COVID-19 dataset. Here we treat different public data resources as different sites. The label distribution across different sites (i.e., public data resources) is heterogeneous naturally.}
\label{fig:site}  
\end{figure}

Once the data are prepared, we employ fully unsupervised and semi-supervised GMMs to analyze the dataset.
We randomly select 70\% of the whole sample for model training. The remaining 30\% are used for validation. 
Among the 70\% training data, 20\% are labeled for semi-supervised learning. 
To better mimic a real decentralized clinical study, we are then inspired to treat different public data resources as different sites \citep{yan2023label}. 
This leads to a total of eight naturally defined sites. 
Moreover, the label distribution across different sites (i.e., public data resources) is heterogeneous naturally.
The label distribution of the dataset is shown in Figure \ref{fig:site}.
The network structure and momentum parameter settings are consistent with those described in Section \ref{sec:competing}. 
The detailed implementation is given in Web Appendix C.2.

\begin{figure}[!h]
\centering
\includegraphics[width=5in]{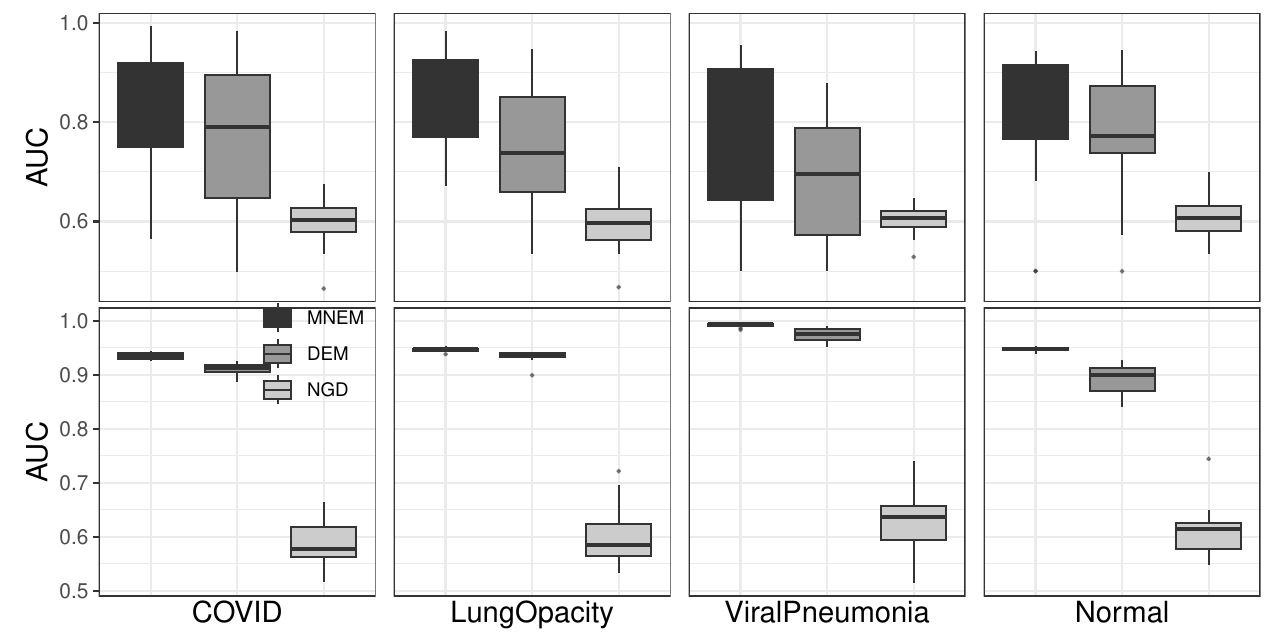} 
\caption{The mean AUC values obtained by different estimators are reported. The upper panel represents the fully unsupervised GMM model, whereas the bottom panel shows the results for its counterparts with partial labeling. Here, we fix the network structure to a circle-type structure.}
\label{fig:boxplot}
\end{figure}

We compute AUC values to evaluate the performance of the proposed methods. For the $k$th class on the $m$th site in the $s$th replicate, define a binary working response as $Y_{ik} = \mathbb{I}(Y_i=k)$. Note that $\mathbb{P}(Y_{ik} = 1 | \theta_0,X_i) = \pi_{ik},$, which is a quantity that depends on $\theta_0.$ We next replace the true parameter $\theta_0$ with one particular estimator (e.g., the semi-MNEM). This leads to an estimate of $\pi_{ik}$, denoted by $\hat{\pi}_{ik}$. Then, an AUC value can be calculated on the basis of $\{(Y_{ik}, \hat{\pi}_{ik}): i \in \mathcal{S}_m\}$. This yields site-specific AUC values for each of the $M$ sites. These AUC values are then averaged to obtain an overall AUC value. The experiment is replicated $20$ times, resulting in a total of
$20$ overall AUC values. 
They are then box-plotted in Figure \ref{fig:boxplot}.

From Figure \ref{fig:boxplot}, we obtain several interesting findings. 
For the unsupervised setting, the MNEM method outperforms the other two estimators with the highest AUC values, while the NGD method has the worst performance. 
In the semi-supervised setting, all methods show significant improvement compared with their unsupervised counterparts. 
For the category \textit{COVID}, the AUC value of MNEM increases from 90\% in the unsupervised setting to 93\% in the semi-supervised setting. Moreover, MNEM remains the best method. In contrast, those of DEM and NGD are 91\% and 58\%, respectively. 
These results are consistent with our theoretical conclusions and highlight the competitive advantage of the proposed method.


\section{CONCLUDING REMARKS} 
\label{sec:discussion}

We conclude by discussing several topics for future study. 
First, the assumed setting may not fully reflect real DFL, which often involves dynamic networks and complex heterogeneities \citep{ye2023heterogeneous}. This limits the generalizability of the proposed MNEM method to some extent. 
Developing novel remedies, such as adaptive network reweighting and domain priors, is therefore promising. 
Lastly, the number of components (i.e., $K$) is assumed known. How to practically determine $K$ in a decentralized way remains an important open problem \citep{kasahara2015testing}. As a preliminary attempt, we developed a decentralized Bayesian information criterion (DBIC); details on the algorithm and simulation are given in Web Appendix C.7. However, its rigorous theory remains to be established, which is another important direction.
\vspace*{-8pt}
\backmatter

\section*{Acknowledgements}

The authors thank the editor, associate editor, and two referees for their valuable comments and suggestions. Xuetong Li and Shuyuan Wu contributed equally as co-first authors. Shuyuan Wu is the corresponding author. 
Xuetong Li's research is partially supported by theChina Postdoctoral Science Foundation (No. 2025M783133). 
Shuyuan Wu's research is partially supported by the National Natural Science Foundation of China (No. 12401392) and China Postdoctoral Science Foundation (No. 2024M751929, No. 2024T170540).
Hansheng Wang's research is partially supported by National Natural Science Foundation of China (72495123, 12271012). 
\vspace*{-8pt}
\section*{SUPPLEMENTARY MATERIALS}

Supplementary material is available at Biometrics online. Web Appendices A--C, referenced in Sections~\ref{sec:methodology}--\ref{sec:simu}, and the code used to reproduce the numerical results are available with this paper on the Biometrics website at Oxford Academic.

\vspace*{-8pt}
\section*{DATA AVAILABILITY}

The data supporting the findings of this paper are publicly available at \url{https://www.kaggle.com/tawsifurrahman/datasets}. 

\vspace*{-8pt}



\bibliographystyle{biom} 
\bibliography{ref}

\label{lastpage}
\end{document}